\documentclass[pdflatex,sn-mathphys-num]{sn-jnl}

\usepackage{graphicx}%
\usepackage{multirow}%
\usepackage{amsmath,amssymb,amsfonts}%
\usepackage{amsthm}%
\usepackage{mathrsfs}%
\usepackage[title]{appendix}%
\usepackage{xcolor}%
\usepackage{textcomp}%
\usepackage{manyfoot}%
\usepackage{booktabs}%
\usepackage{listings}%
\usepackage[ruled,vlined]{algorithm2e} 
\usepackage{tabularx} 

\theoremstyle{thmstyleone}%
\theoremstyle{thmstyletwo}%

\theoremstyle{thmstylethree}%

\begin{document}

\title[Article Title]{FedMust: Semi-supervised Multi-task Student-Teacher Federated Learning for Multi-organ CT Segmentation}

\author*[1]{\fnm{Ashkan} \sur{Moradi}}
\author[1]{\fnm{Bendik} \sur{Skarre Abrahamsen}}
\author[1,2]{\fnm{Mattijs} \sur{Elschot}}

\affil*[1]{\orgdiv{Department of Circulation and Medical Imaging}, \orgname{Norwegian University of Science and Technology}, \orgaddress{\city{Trondheim}, \country{Norway}}
\email{\{ashkan.moradi, bendik.s.abrahamsen, mattijs.elschot\}@ntnu.no}}
\affil[2]{\orgdiv{Central Staff at St. Olavs Hospital}, \orgname{Trondheim University Hospital}, \orgaddress{\city{Trondheim}, \country{Norway}}}

\abstract{
\textbf{Purpose:} Multi-organ segmentation using deep learning requires large amounts of annotated patient data, which are often unavailable at individual institutions. Privacy constraints further limit data sharing, while the labor-intensive nature of annotation leaves substantial amounts of local data unlabeled. We propose a flexible semi-supervised federated multi-task student–teacher (FedMust) framework that leverages both labeled and unlabeled data across participating sites within a federated learning (FL) setting for multi-organ segmentation.

\textbf{Methods:} At each communication round, clients with labels for the same task form a federation to train an aggregated teacher model. These teacher models generate task-specific features for all data at each client. Subsequently, all clients form another federation to train a multi-task student model comprising a shared encoder and task-specific decoders that replicate the teacher-generated features across segmentation tasks. The aggregated student model is then used to update the local teachers for the next training round.

\textbf{Results:} Extensive experiments demonstrated that FedMust outperformed local and federated single-organ models, yielding an average performance gain of approximately 13\% across clients ($p < 0.01$). The experiments also demonstrated the impact of multi-task learning and unlabeled data and the applicability of the framework in relaxing labeled-data requirements for client participation.

\textbf{Conclusion:} FedMust provides a flexible approach for federated multi-organ segmentation by leveraging distributed labeled and unlabeled data while relaxing the requirements for labeled data for client participation in the federation.
}


\maketitle

\section{Introduction}

The use of deep learning techniques for organ segmentation has improved the accuracy and robustness of multi-organ segmentation models; however, such methods typically require substantial amounts of annotated data~\cite{fu2021review,tajbakhsh2020embracing}. Data annotation is time-consuming and labor-intensive, resulting in only a small portion of existing data in institutions being annotated and utilized for segmentation. Moreover, many institutions lack the resources required for multi-organ annotation, while their available labels are fragmented across different clinical fields. In such cases, data sharing among institutions is also prevented due to patient data privacy constraints. To mitigate data scarcity and privacy concerns, federated learning (FL) has emerged as a promising solution by enabling collaborative model training on larger, more diverse datasets without sharing raw data. In FL, a group of clients train models on their local data and share only model parameters or updates with a trusted server, which aggregates them to obtain a global model with improved generalizability~\cite{Communication2017McMahan}. Consequently, FL has become a viable approach in digital health~\cite{rieke2020future}, with applications in cancer detection~\cite{pati2022federated,sheller2019multi,moradi2025optimizing,moradi2024federated,moradi2025beyond} and medical image segmentation~\cite{xu2021federated,sheller2020federated}. Despite its potential, FL faces challenges in real-world implementation~\cite{kairouz2021advances,li2020federated}, particularly in healthcare settings where data and model heterogeneity arise from variations in data volume, acquisition protocols, imaging modalities, and data distributions across clients~\cite{ma2024model,sheller2020federated}.

Most FL segmentation approaches require fully annotated datasets for client participation; however, in real-world scenarios, such data are not available. To address this, semi-supervised FL techniques have been proposed to enable participation and leverage partially annotated data. Semi-supervised organ segmentation has been studied in both centralized and federated settings~\cite{huang2020multi,fang2020multi}; however, multi-task approaches that address inconsistent annotations remain limited~\cite{zhang2021dodnet}. Early foundational work established a baseline for multi-organ and tumor segmentation from non-overlapping labels~\cite{shen2022joint}, but demonstrated that the aggregated global model can drift from local models due to task heterogeneity. To mitigate such knowledge conflicts, various frameworks have been proposed, including loss function design~\cite{shi2021marginal}, double-teacher training with knowledge distillation~\cite{zhan2023fed}, adaptive server aggregation to penalize clients with higher conflicts~\cite{du2025fedrs}, and combinations of such approaches~\cite{wang2023condistfl}. Moreover, architectural designs with task-specific encoders~\cite{xu2023federated} and decoders~\cite{kim2024federated} mitigate knowledge conflicts among tasks; yet, they require prior knowledge of all tasks before training and annotations for at least one task per client, respectively. Combining these approaches can further improve the performance and generalizability of FL under partially labeled data~\cite{deng2025fedsemidg,tolle2025real}. However, many multi-task semi-supervised FL methods still fail when no group of clients shares a common annotated task, and the impact of the typically predominant unlabeled local data remains largely unexplored.

In this work, we propose a flexible, semi-supervised, federated multi-task student-teacher (FedMust) framework that trains a multi-organ segmentation model using a mixture of labeled and unlabeled data from clients. This approach employs a feature-matching technique to extract maximal information from partially labeled and unlabeled client data.  FedMust then trains a student model with a shared encoder and frozen task-specific decoders, thereby placing greater emphasis on learning task-agnostic and generalizable features while preventing deviation from the task-specific teacher models. Iteratively training and updating the teacher and student models in this manner improves the overall multi-organ segmentation performance across clients. The main contributions of this work are summarized as follows:
\begin{itemize}
    \item We propose a flexible, semi-supervised federated multi-task learning framework for multi-organ segmentation that operates through the iterative training of student and teacher models.
    \item We introduce a feature-matching strategy that distills information from partially labeled and unlabeled local data to enhance segmentation performance, along with a single-encoder, multi-decoder student model that generalizes segmentation features while retaining task-specific performance.
    \item We conduct extensive experiments to investigate the impact of multi-task learning and the inclusion of unlabeled data from other clients. Evaluation on two unseen test sets demonstrates improved performance compared with models trained in isolation.
    \item We present an ablation study to further investigate the impact of incorporating unlabeled data from the local clients themselves and the contribution of an additional client without annotated data to the FL process.
\end{itemize}

The remainder of the paper is structured as follows: Section~\ref{sec_method} outlines the proposed methodology, Section~\ref{sec_exp_res} presents the designed experiments and results, and Section~\ref{sec_abla} presents the ablation studies. The findings are discussed in Section~\ref{sec_discu}, and concluding remarks are provided in Section~\ref{sec_conclu}.

\section{Methodology}\label{sec_method}

Consider a set of clients that aim to collaboratively train a model for multi-task abdominal organ segmentation over a known set of tasks. In this setting, each client possesses only a small amount of annotated data for a subset of the tasks, while the majority of its data remains unlabeled. Restricted by privacy constraints, the clients cannot directly share their data to train a centralized model. Furthermore, since the sets of labeled tasks available across clients do not necessarily overlap, traditional FL cannot be directly employed. Training models independently at each client is also limited by the scarcity of annotations required to perform segmentation across the entire set of tasks.

Under this setting, we denote the set of clients participating in the federation by $\mathcal{C}$, with each client represented by $i \in \mathcal{C}$. The segmentation of each abdominal organ is considered a separate task, denoted by $k \in \mathcal{K}$, where $\mathcal{K}$ represents the set of all segmentation tasks. Each client possesses partially labeled data, i.e., annotations are available for only a subset of the tasks. We denote the subset of tasks for which annotations are available at client~$i$ by $\mathcal{K}_i \subseteq \mathcal{K}$. Furthermore, $\mathcal{N}_i^k$ and $\bar{\mathcal{N}}_i^k$ denote the sets of labeled and unlabeled data available at client~$i$ for task~$k$, respectively. The set of clients with annotated data for task~$k$ is denoted by $\mathcal{C}_k$, and these clients are eligible to participate in the federated training of the corresponding teacher model.

At federated round~$r$, each client $i \in \mathcal{C}_k$ constructs a teacher model for task $k \in \mathcal{K}i$ using a standard 3D~U-Net architecture. The teacher model comprises two sets of trainable parameters, $\mathbf{W}_{\text{T}_i}^{\text{En}}(r,k)$ and $\mathbf{W}_{\text{T}_i}^{\text{De}}(r,k)$, corresponding to the encoder and decoder, respectively. Each teacher model is then trained for $E$ local epochs using the corresponding labeled data $\mathcal{N}_i^k$. The resulting teacher models are then aggregated at the server using FedAvg~\cite{Communication2017McMahan} as 
\begin{equation}
    \label{tea_aggr}
    \begin{aligned}
        \mathbf{W}_{\text{T}_{\text{FL}}}^{\text{En}}(r,k) = \sum_{i \in \mathcal{C}_k} \frac{|\mathcal{N}_i^k|}{\sum_{j \in \mathcal{C}_k}|\mathcal{N}_j^k|}~ \mathbf{W}_{\text{T}_{i}}^{\text{En}}(r,k), \forall ~k \in \mathcal{K},\\
        \mathbf{W}_{\text{T}_{\text{FL}}}^{\text{De}}(r,k) = \sum_{i \in \mathcal{C}_k} \frac{|\mathcal{N}_i^k|}{\sum_{j \in \mathcal{C}_k}|\mathcal{N}_j^k|}~ \mathbf{W}_{\text{T}_{i}}^{\text{De}}(r,k), \forall ~k \in \mathcal{K}.
    \end{aligned}
\end{equation}
The aggregated parameter sets $(\mathbf{W}_{\text{T}_{\text{FL}}}^{\text{En}}(r,k), \mathbf{W}_{\text{T}_{\text{FL}}}^{\text{De}}(r,k)) \forall~k \in \mathcal{K}$ are then returned to the clients and subsequently used to generate features from all available data, $\mathcal{N}_i^k \cup \bar{\mathcal{N}}_i^k$, for all tasks $k \in \mathcal{K}$ and clients $i \in \mathcal{C}$.

\begin{figure*}[t]
    \centering
    \includegraphics[width=\textwidth]{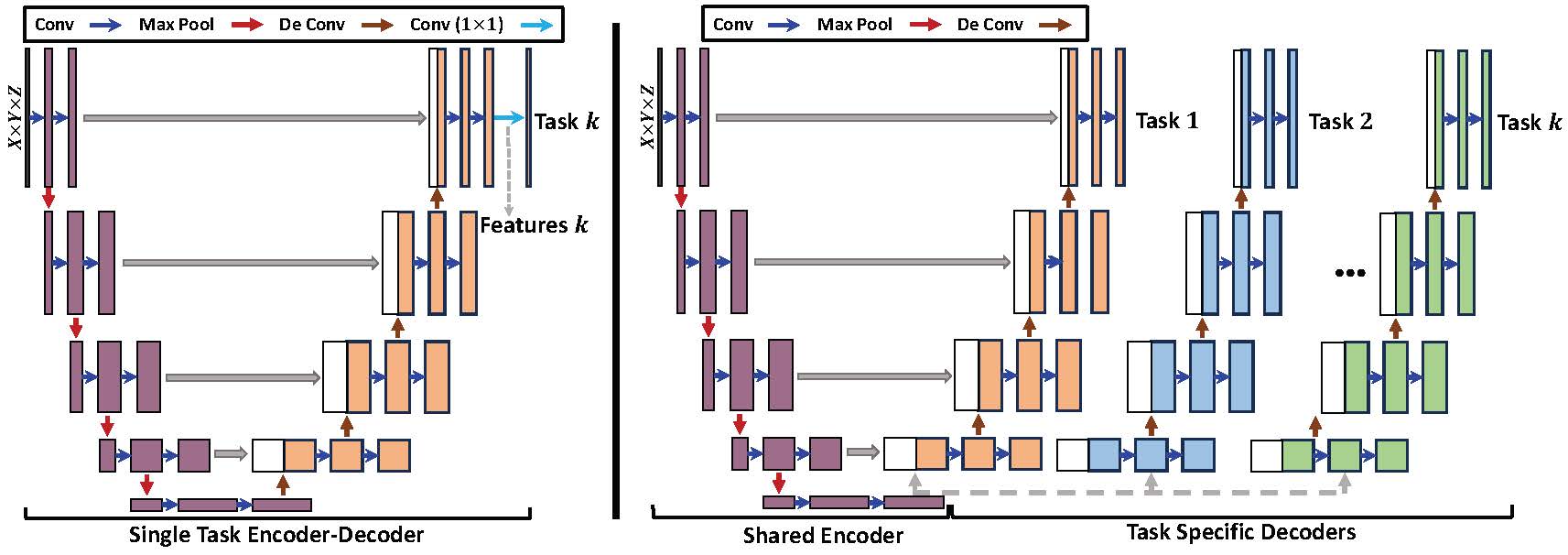}
    \caption{Model architectures of the single-task 3D U-Net teacher (left) and the multi-task 3D U-Net student with task-specific decoders (right).} \label{stu_tea_model}
\end{figure*}

The teacher features are defined as the outputs of the teacher model immediately before its final fully connected layer. For further clarification, see the teacher model illustrated in Figure~\ref{stu_tea_model}. Following feature generation, student training is initiated, where each client $i \in \mathcal{C}$ constructs a student model as a multi-decoder 3D~U-Net architecture. The student model consists of a shared encoder and multiple task-specific decoders, with each decoder corresponding to a specific task. Unlike the teacher models, the student decoders exclude the final fully connected layer. The architectures of the teacher and proposed student models are illustrated in Figure~\ref{stu_tea_model}.

The student model is defined by two sets of parameters: the shared encoder parameters $\mathbf{W}_{\text{S}_i}^{\text{En}}(r)$ and the task-specific decoder parameters $\mathbf{W}_{\text{S}_i}^{\text{De}}(r, k)$. For each task~$k$, the corresponding aggregated teacher decoder parameters are transferred to the student decoder, $\mathbf{W}_{\text{T}_{\text{FL}}}^{\text{De}}(r, k) \rightarrow \mathbf{W}_{\text{S}_i}^{\text{De}}(r, k)$, and subsequently frozen during student training. The student model is then trained by optimizing the shared encoder to minimize the mean squared error (MSE) between the features generated by the student and those produced by the corresponding teacher models. This encourages the shared student encoder to learn more generalizable features, while the frozen task-specific decoders constrain deviation from the teacher models. For each student, the overall MSE loss is defined as $\mathcal{L}^{(\text{St})} = \sum_{k \in \mathcal{K}} \mathcal{L}_k$, where the task-specific MSE loss $\mathcal{L}_k$ is defined as

\begin{equation}
    \begin{aligned}
        \mathcal{L}_k
        &= \left\| F_k^{(\mathrm{Te})} - F_k^{(\mathrm{St})} \right\|_F^2 \\
        &= \frac{1}{XYZ}
        \sum_{x=1}^{X}\sum_{y=1}^{Y}\sum_{z=1}^{Z}
        \left(
        f_k^{(\mathrm{Te})}(x,y,z)
        -
        f_k^{(\mathrm{St})}(x,y,z)
        \right)^2 .
    \end{aligned}
\end{equation}
with {$\| \cdot \|_F$} denoting the Frobenius norm, and 
{$F_k^{(\text{Te})}, F_k^{(\text{St})} \in \mathbb{R}^{X\times Y\times Z}$}
representing the teacher and student features for task $k$, respectively. Here, {$X$} and {$Y$} denote the spatial dimensions, and {$Z$} is the feature depth. The feature voxel of task $k$ at spatial location $(x, y, z)$ is denoted by {$f^{(\text{Te})}_k(x,y,z)$} and {$f^{(\text{St})}_k(x,y,z)$} for the teacher and the student, respectively. The MSE loss then enforces {$F_k^{(\text{Te})} \approx F_k^{(\text{St})}$}, encouraging the student representation to closely match that of the corresponding teacher. 

The student models are trained for $E$ local epochs, after which the encoder parameters are sent to the server for aggregation. The trained encoder parameters are then aggregated at the server using FedAvg~\cite{Communication2017McMahan} as
\begin{equation}
    \label{stu_aggr}
    \mathbf{W}_{\text{S}_{\text{FL}}}^{\text{En}}(r) = \sum_{i \in \mathcal{C}} \frac{|\mathcal{N}_i^k \cup \bar{\mathcal{N}}_i^k|}{\sum_{j \in \mathcal{C}}|\mathcal{N}_j^k \cup \bar{\mathcal{N}}_j^k|}~ \mathbf{W}_{\text{S}_i}^{\text{En}}(r),
\end{equation}
and subsequently returned to the clients to update the teacher encoder parameters, $\mathbf{W}_{\text{S}_{\text{FL}}}^{\text{En}}(r) \rightarrow \mathbf{W}_{\text{T}_i}^{\text{En}}(r, k)$ for all clients $i \in \mathcal{C}$ and $k \in \mathcal{K}$. The updated teacher models are subsequently used to initiate the next round of training, and this iterative process continues for a predefined number of federated rounds. A summary of the proposed algorithm for each client $i \in \mathcal{C}$ is provided in Algorithm~\ref{alg_stu_tea}.

\begin{algorithm}[t]
    \SetAlgoLined
    \caption{FedMust: Multi-task Student-Teacher Federated Learning}
    \label{alg_stu_tea}
    \SetAlgoLined
    \textbf{Initialize:} {$\mathbf{W}_{\text{T}_i}^{\text{En}}(0,k), \mathbf{W}_{\text{T}_i}^{\text{De}}(0,k)$} for all {$k \in \mathcal{K}_i$}\;
    \For{global round {$r = 0, \cdots, R-1$}}{
        \For{each task {$k \in \mathcal{K}_i$}}{
            \For{each client {$i \in \mathcal{C}_k$} \text{in parallel}}{
                Update {$\mathbf{W}_{\text{T}_{i}}^{\text{En}}(r+1,k), \mathbf{W}_{\text{T}_{i}}^{\text{De}}(r+1,k)$} by \\{\textsf{Teacher-Train}($\mathcal{N}_i^k$)\;}}
            Aggregate teacher models: \\
            {$\mathbf{W}_{\text{T}_{\text{FL}}}^{\text{En}}(r+1,k) \xleftarrow{\text{AVG}} \{\mathbf{W}_{\text{T}_{i}}^{\text{En}}(r+1,k)\}_{i \in \mathcal{C}_k}$} \\
            {$\mathbf{W}_{\text{T}_{\text{FL}}}^{\text{De}}(r+1,k) \xleftarrow{\text{AVG}} \{\mathbf{W}_{\text{T}_{i}}^{\text{De}}(r+1,k)\}_{i \in \mathcal{C}_k}$}\;}
        Generate teacher-features for all {$k \in \mathcal{K}$} and {$i \in \mathcal{C}$}\;
        Update students for all {$i \in \mathcal{C}$ and $k \in \mathcal{K}$}: \\ 
        {$\mathbf{W}_{\text{S}_{i}}^{\text{De}}(r+1,k) \leftarrow \mathbf{W}_{\text{T}_{\text{FL}}}^{\text{De}}(r+1,k)$}\;
        \For{each client {$i \in \mathcal{C}$} \text{in parallel}}{
            Update {$\mathbf{W}_{\text{S}_{i}}^{\text{En}}(r+1)$} by \\{\textsf{Student-Train}($\mathcal{N}_i^k \cup \bar{\mathcal{N}}_i^k$})\;}
        Aggregate student models: 
        {$\mathbf{W}_{\text{S}_{\text{FL}}}^{\text{En}}(r+1) \xleftarrow{\text{AVG}} \{\mathbf{W}_{\text{S}_{i}}^{\text{En}}(r+1)\}_{i \in \mathcal{C}}$}\;
        Update teachers for all $k \in \mathcal{K}$ and $i \in \mathcal{C}_k$: \\ 
        {$\mathbf{W}_{\text{T}_{i}}^{\text{En}}(r+1,k) \leftarrow \mathbf{W}_{\text{S}_{\text{FL}}}^{\text{En}}(r+1)$}\;}
    \textbf{Output:} {$\mathbf{W}_{\text{S}_{\text{FL}}}^{\text{En}}(R), \mathbf{W}_{\text{S}_{\text{FL}}}^{\text{De}}(R,k)$} for all {$k \in \mathcal{K}$}\;
\end{algorithm}

\section{Experimental Results}\label{sec_exp_res}

For the experimental setup, we consider a scenario involving four institutions that aim to collaborate in a privacy-preserving manner to perform multi-organ abdominal CT segmentation. The scenario comprises four clients, each possessing annotations for a single-organ CT dataset. Specifically, clients C1, C3, and C4 possess CT images and corresponding annotations for 131 liver, 281 pancreas, and 41 spleen scans from the Medical Segmentation Decathlon challenge~\cite{antonelli2022medical}, respectively. Client C2 holds 210 labeled kidney CT images from the KiTS19 challenge~\cite{heller2019kits19}. The annotations for the left and right kidneys are grouped together under a single annotation label, and kidney segmentation in this study refers to the segmentation of both kidneys. The available data are partitioned using five-fold cross-validation, with 80\% used for training and 20\% for validation in each fold. Table~\ref{annotation_table} summarizes data and annotation availability for each task and client.

\begin{table*}[t]
    \centering
    \caption{Data distribution for training, validation, and testing across clients and the independent test set. Label availability for each task (liver, kidney, pancreas, and spleen) is indicated by $\boldsymbol{\checkmark}$, while label absence is indicated by $\boldsymbol{\times}$.}
    \label{annotation_table}
    \resizebox{\textwidth}{!}{%
    \begin{tabular}{>{\raggedright\arraybackslash}p{3.2cm}>{\centering\arraybackslash}p{4cm}*{4}{>{\centering\arraybackslash}p{1.5cm}}}
        \toprule 
        \textbf{Client} & \textbf{Data/train/val/test} & \textbf{Liver} & \textbf{Kidney} & \textbf{Pancreas} & \textbf{Spleen}\\ 
        \midrule 
        \textbf{C1} (MSD~\cite{antonelli2022medical}) & 131/105/26/0 & $\boldsymbol{\checkmark}$ & $\boldsymbol{\times}$ & $\boldsymbol{\times}$ & $\boldsymbol{\times}$ \\ 
        \textbf{C2} (KiTS19~\cite{heller2019kits19}) & 210/168/42/0 & $\boldsymbol{\times}$ & $\boldsymbol{\checkmark}$ & $\boldsymbol{\times}$ & $\boldsymbol{\times}$\\ 
        \textbf{C3} (MSD~\cite{antonelli2022medical}) & 281/225/56/0 & $\boldsymbol{\times}$ & $\boldsymbol{\times}$ & $\boldsymbol{\checkmark}$ & $\boldsymbol{\times}$ \\ 
        \textbf{C4} (MSD~\cite{antonelli2022medical}) & 41/33/8/0 & $\boldsymbol{\times}$ & $\boldsymbol{\times}$ & $\boldsymbol{\times}$ & $\boldsymbol{\checkmark}$ \\ 
        \textbf{AMOS22~\cite{ji2022amos}} & 276/0/0/276 & $\boldsymbol{\checkmark}$ & $\boldsymbol{\checkmark}$ & $\boldsymbol{\checkmark}$ & $\boldsymbol{\checkmark}$\\
        \textbf{TotalSegmentator~\cite{wasserthal2023totalsegmentator}} & 626/0/0/626 & $\boldsymbol{\checkmark}$ & $\boldsymbol{\checkmark}$ & $\boldsymbol{\checkmark}$ & $\boldsymbol{\checkmark}$\\
        \bottomrule 
    \end{tabular}}
\end{table*}

The objective of each institution is to train a model capable of segmenting all four organs. Due to privacy restrictions, centralizing the data is not feasible, while traditional FL approaches are not applicable when no group of clients shares annotations for the same task. The proposed FedMust algorithm addresses this limitation by enabling the development of a multi-task organ segmentation model across such clients. The resulting model is then compared with conventional FL and client-specific models trained in isolation on local data. The trained models are evaluated on two independent test sets containing annotations for all tasks: AMOS~\cite{ji2022amos} (276 cases) and TotalSegmentator~\cite{wasserthal2023totalsegmentator} (626 cases). For both datasets, only cases with non-empty annotations for all four organs are included in the test set. Evaluation on these out-of-distribution test sets provides insight into the generalizability of the trained models and their ability to learn representations beyond client-specific features.

\begin{figure*}[t]
    \centering
    \includegraphics[width=\textwidth]{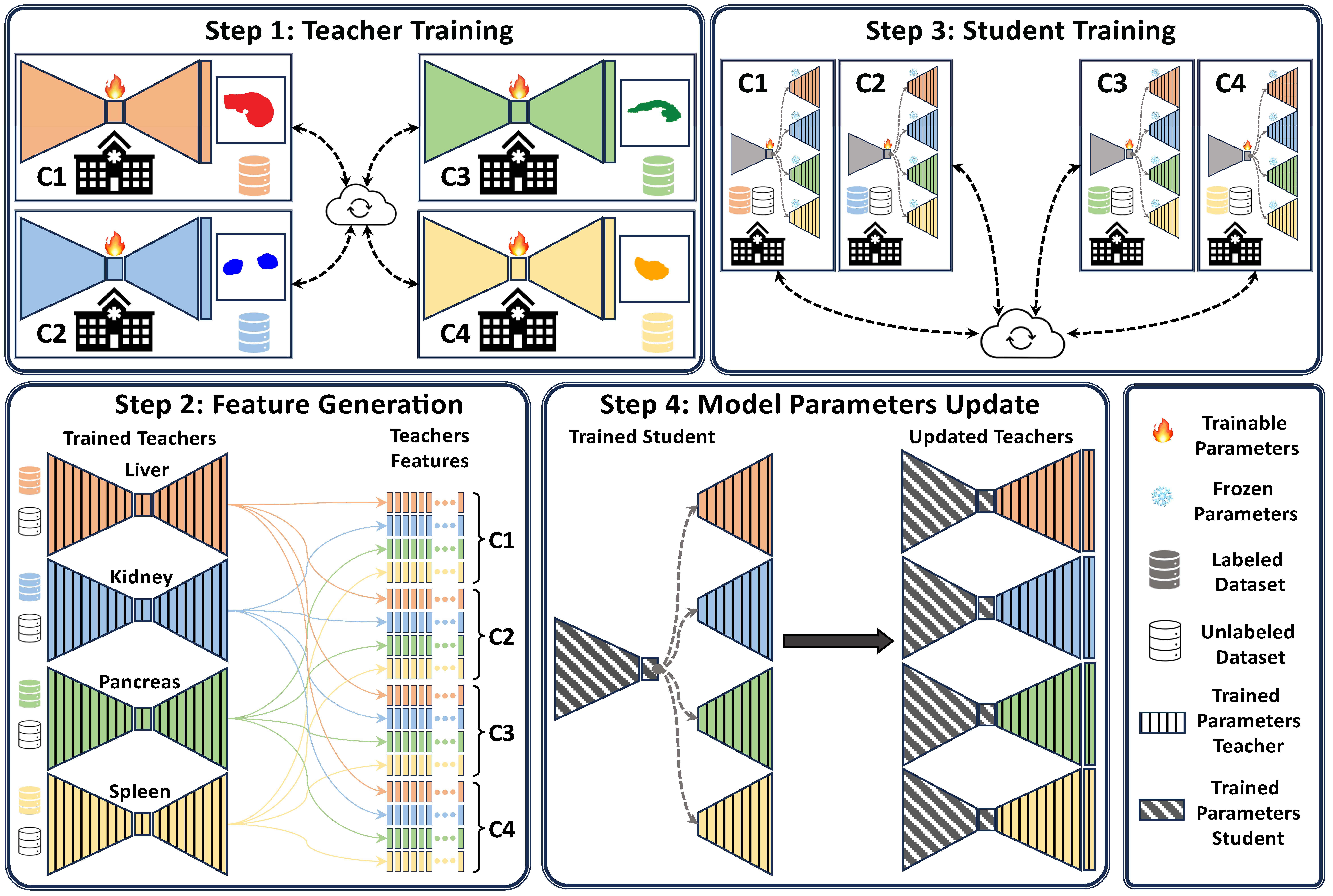}
    \caption{
    Overview of the proposed multi-task semi-supervised student–teacher training framework in an FL network with four clients. Each client has labeled data for a single organ (C1: liver; C2: kidney; C3: pancreas; C4: spleen). The proposed training procedure is illustrated in four steps.} \label{dynamic}
\end{figure*}

Figure~\ref{dynamic} summarizes the workflow of the proposed FedMust algorithm for the described scenario. The process begins in Step~1 by training a 3D U-Net model on the available labeled data at each client. This is followed by Step~2, in which the trained teacher models are used to generate features for all available labeled and unlabeled data across tasks. In Step~3, a 3D U-Net-based multi-decoder student model is then trained to replicate these features while its decoders remain frozen. The hypothesis is that incorporating features from unlabeled data improves the overall performance of the organ segmentation models, while freezing the student decoders during training prevents the model from drifting away from the task-specific teachers and allows the shared encoder to learn task-agnostic features. Finally, in Step~4, the teacher models are updated, and another round of FL is initiated.

\subsection{Pre-processing and Implementation Details}
All experiments were implemented in Python~3.10 and conducted on an NVIDIA~A40~GPU with 48~GB of memory, using CUDA~12.4 on Ubuntu 20.04.6~LTS. The encoder-decoder models were designed and implemented using MONAI~1.4.0. Prior to model training, the images were pre-processed by resampling to a uniform voxel resolution of $1.5\,\mathrm{mm} \times 1.5\,\mathrm{mm} \times 3.0\,\mathrm{mm}$, cropping to 128 slices of $224 \times 224$ voxels each, and clipping the image intensities to the 1st--99th percentile range, followed by intensity normalization.

During training, data augmentation was performed using random flipping along the first spatial axis and random $90^\circ$ rotations, each applied with a probability of 0.5. The Adam optimizer was used for training with an initial learning rate of 0.01 and linearly reduced to $10^{-5}$. For FL aggregation, FedAvg was used to aggregate the models proportionally to each client’s dataset size. The batch size was set to 4, with each epoch iterating over all mini-batches. For all experiments, the local models were trained for E=300 iterations, which was sufficient for convergence. To ensure a fair comparison, the FL experiments were configured with the same total number of training iterations, using E=30 local iterations per communication round for R=10 communication rounds. Performance was evaluated using the 3D Dice score, with 95\% confidence intervals estimated through bootstrap resampling with a sample size of \scalebox{0.9}{$b = 1000000$}. All reported metrics were computed based on the ensembled performance of the five-fold cross-validated models. The ensembled model generates segmentation predictions by averaging the prediction maps obtained from the five corresponding cross-validated models.

\begin{sidewaystable}
    \caption{Ensembled Dice performance of the models trained using five-fold cross-validation and evaluated on the independent AMOS (top) and TotalSegmentator (bottom) test sets, reported with 95\% confidence intervals. The table illustrates the impact of the proposed model, particularly its ability to leverage unlabeled data from other clients to improve segmentation performance.}
    \label{table_amos_totseg_num_lbl}
    \begin{tabular*}{\textheight}{@{\extracolsep\fill}lllcccccc}
    \toprule
    & & & \multicolumn{6}{c}{\textbf{Number of Labeled Data}} \\
    \cmidrule(lr){4-9}
    \textbf{Test Set} & \textbf{Task} & \textbf{Model} &
    \textbf{10-Labeled} & \textbf{20-Labeled} & \textbf{40-Labeled} &
    \textbf{60-Labeled} & \textbf{80-Labeled} & \textbf{All-Labeled} \\
    \midrule
    
    \multirow{8}{*}{AMOS22}
    & \multirow{2}{*}{Liver} & Local
    & \textbf{0.78}~(0.76,0.81)
    & \textbf{0.84}~(0.83,0.86)
    & \textbf{0.84}~(0.82,0.85)
    & \textbf{0.85}~(0.82,0.88)
    & \textbf{0.86}~(0.84,0.88)
    & 0.87~(0.85,0.88) \\
    
    & & FedMust
    & 0.67~(0.64,0.71)
    & 0.81~(0.79,0.83)
    & \textbf{0.84}~(0.81,0.85)
    & \textbf{0.85}~(0.83,0.86)
    & 0.85~(0.83,0.87)
    & \textbf{0.88}~(0.86,0.89) \\
    \cmidrule(lr){2-9}
    
    & \multirow{2}{*}{Kidney} & Local
    & 0.30~(0.26,0.35)
    & 0.41~(0.37,0.45)
    & 0.48~(0.44,0.53)
    & 0.54~(0.49,0.58)
    & 0.67~(0.63,0.71)
    & 0.78~(0.75,0.81) \\
    
    & & FedMust
    & \textbf{0.49}~(0.45,0.52)
    & \textbf{0.45}~(0.40,0.49)
    & \textbf{0.55}~(0.50,0.59)
    & \textbf{0.64}~(0.60,0.68)
    & \textbf{0.71}~(0.68,0.75)
    & \textbf{0.84}~(0.81,0.86) \\
    \cmidrule(lr){2-9}
    
    & \multirow{2}{*}{Pancreas} & Local
    & 0.05~(0.04,0.06)
    & 0.17~(0.15,0.19)
    & \textbf{0.24}~(0.21,0.26)
    & 0.28~(0.26,0.31)
    & 0.27~(0.24,0.30)
    & 0.45~(0.41,0.48) \\
    
    & & FedMust
    & \textbf{0.11}~(0.09,0.12)
    & \textbf{0.18}~(0.15,0.20)
    & 0.21~(0.17,0.23)
    & \textbf{0.32}~(0.30,0.35)
    & \textbf{0.33}~(0.30,0.37)
    & \textbf{0.52}~(0.49,0.55) \\
    \cmidrule(lr){2-9}
    
    & \multirow{2}{*}{Spleen} & Local
    & \textbf{0.62}~(0.58,0.66)
    & 0.59~(0.55,0.63)
    & \textbf{0.67}~(0.63,0.71)
    & 0.63~(0.59,0.67)
    & 0.63~(0.59,0.67)
    & 0.63~(0.59,0.67) \\
    
    & & FedMust
    & 0.61~(0.57,0.65)
    & \textbf{0.66}~(0.62,0.70)
    & \textbf{0.67}~(0.63,0.71)
    & \textbf{0.67}~(0.63,0.71)
    & \textbf{0.67}~(0.63,0.71)
    & \textbf{0.68}~(0.64,0.72) \\
    
    \midrule
    \multicolumn{3}{c}{\textbf{Avg improvement \%}}
    & \textbf{41.9}\%
    & \textbf{6.0}\%
    & \textbf{1}\%
    & \textbf{9.5}\%
    & \textbf{8.3}\%
    & \textbf{8.1}\% \\
    
    \midrule
    
    \multirow{8}{*}{TotalSeg}
    & \multirow{2}{*}{Liver} & Local
    & \textbf{0.76}~(0.74,0.78)
    & \textbf{0.85}~(0.84,0.86)
    & 0.85~(0.84,0.87)
    & \textbf{0.87}~(0.86,0.88)
    & 0.87~(0.85,0.88)
    & 0.91~(0.90,0.92) \\
    
    & & FedMust
    & 0.72~(0.70,0.74)
    & \textbf{0.85}~(0.84,0.86)
    & \textbf{0.86}~(0.85,0.87)
    & \textbf{0.87}~(0.86,0.89)
    & \textbf{0.89}~(0.88,0.90)
    & \textbf{0.92}~(0.91,0.93) \\
    \cmidrule(lr){2-9}
    
    & \multirow{2}{*}{Kidney} & Local
    & 0.42~(0.38,0.45)
    & 0.47~(0.44,0.50)
    & 0.53~(0.49,0.56)
    & \textbf{0.57}~(0.54,0.60)
    & \textbf{0.62}~(0.59,0.65)
    & 0.71~(0.68,0.73) \\
    
    & & FedMust
    & \textbf{0.51}~(0.48,0.54)
    & \textbf{0.53}~(0.50,0.56)
    & \textbf{0.57}~(0.54,0.60)
    & \textbf{0.57}~(0.54,0.60)
    & 0.60~(0.57,0.63)
    & \textbf{0.73}~(0.70,0.75) \\
    \cmidrule(lr){2-9}
    
    & \multirow{2}{*}{Pancreas} & Local
    & 0.07~(0.06,0.08)
    & 0.17~(0.16,0.19)
    & 0.26~(0.25,0.28)
    & 0.33~(0.31,0.35)
    & 0.37~(0.35,0.39)
    & 0.51~(0.48,0.53) \\
    
    & & FedMust
    & \textbf{0.13}~(0.12,0.15)
    & \textbf{0.21}~(0.19,0.22)
    & \textbf{0.32}~(0.30,0.34)
    & \textbf{0.41}~(0.39,0.43)
    & \textbf{0.42}~(0.39,0.44)
    & \textbf{0.57}~(0.55,0.60) \\
    \cmidrule(lr){2-9}
    
    & \multirow{2}{*}{Spleen} & Local
    & 0.50~(0.47,0.52)
    & 0.54~(0.51,0.57)
    & 0.58~(0.55,0.61)
    & 0.57~(0.55,0.60)
    & 0.57~(0.55,0.60)
    & 0.57~(0.55,0.60) \\
    
    & & FedMust
    & \textbf{0.63}~(0.61,0.66)
    & \textbf{0.64}~(0.62,0.67)
    & \textbf{0.69}~(0.66,0.71)
    & \textbf{0.68}~(0.65,0.71)
    & \textbf{0.69}~(0.67,0.72)
    & \textbf{0.70}~(0.67,0.72) \\
    
    \midrule
    \multicolumn{3}{c}{\textbf{Avg improvement \%}}
    & \textbf{32.0}\%
    & \textbf{13.7}\%
    & \textbf{12.7}\%
    & \textbf{10.9}\%
    & \textbf{8.4}\%
    & \textbf{9.6}\% \\
    
    \botrule
    \end{tabular*}
\end{sidewaystable}

\subsection{Multi-tasking and Partially Labeled Data}\label{sec_num_lbl} 
As illustrated in Fig.~\ref{dynamic}, clients possess partially labeled data, meaning that each client has labels for only a subset of the available tasks. In this section, we investigate the impact of utilizing unlabeled data from other clients, through the multi-task training of FedMust, on improving the segmentation performance for specific tasks. The proposed FedMust algorithm is employed to train a model that provides segmentation capabilities across all clients and tasks. To evaluate its performance, we compare the Dice scores for organ segmentation with those obtained by local models trained independently at each client, as conventional FL fails in this setting. To further highlight the impact of the amount of unlabeled data from other clients on performance, we repeated the experiment with varying amounts of labeled data available at each client. This allows us to isolate the effect of incorporating additional unlabeled task data from other clients during student training on the final performance. Thus, the performance gains achieved by FedMust as a function of the amount of local labeled data and unlabeled data from other clients are obtained.

Table~\ref{table_amos_totseg_num_lbl} reports the ensembled performance of the 5-fold cross-validated models, together with their 95\% confidence intervals. As shown, FedMust outperforms the local models, particularly when the local models exhibit poor performance. The observed performance gains of FedMust indicate that multi-task segmentation, through the incorporation of unlabeled data from other clients into the federation during student model training, further refines the feature extraction capability of the student encoder. This is more clearly demonstrated for the C4:~Spleen client, where, due to the limited number of labeled samples, the same 41 samples are used in the last three experiments. As a result, the trained local models achieve identical performance, while FedMust continues to improve by leveraging additional unlabeled data from other clients. Incorporating unlabeled data from other clients results in a more generalizable shared multi-task encoder through student training. 

\begin{figure*}[t]
    \centering
    \includegraphics[width=\textwidth]{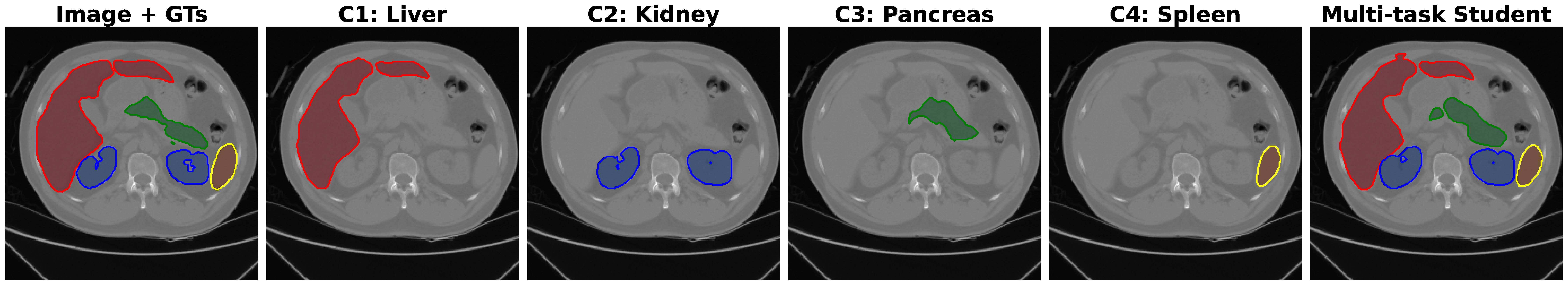}\vspace{-1mm}
    \includegraphics[width=\textwidth]{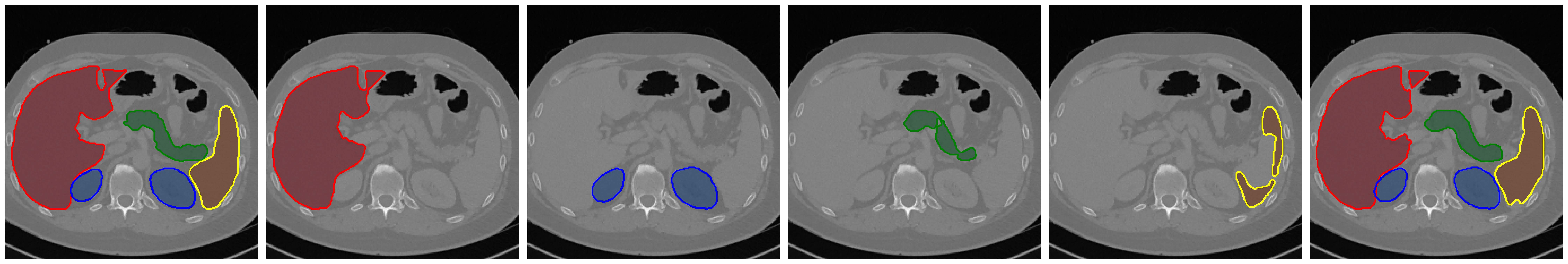}
    \caption{Qualitative performance on representative test cases from the AMOS and TotalSegmentator datasets. The first column shows the CT images with their corresponding ground-truth annotations. The subsequent client columns show predictions from the local models, which are limited to the tasks annotated at each client. The last column shows FedMust predictions for all tasks, demonstrating improved segmentation performance across all tasks.}\label{qualitative}
\end{figure*}

The qualitative performance of the proposed FedMust algorithm for multi-organ segmentation is illustrated in Figure~\ref{qualitative}. As illustrated, clients trained in isolation can only provide segmentations for tasks for which labeled data are locally available, whereas FedMust enables multi-organ segmentation with improved performance across all tasks. This improvement can be observed by comparing the multi-organ segmentation predictions with the corresponding ground-truth annotations.

\section{Ablation Studies}\label{sec_abla}
In this section, we conduct ablation studies to further investigate key aspects of the proposed FedMust framework. Specifically, we examine the utilization of unlabeled data available at participating clients and the possibility of including clients without annotations in the federation.

\subsection{Utilization of Unlabeled Client Data}\label{local_unlabled_sec}
\begin{table*}[t]
    \centering
    \caption{Ensembled Dice performance of the models trained using five-fold cross-validation and evaluated on the independent AMOS (top) and TotalSegmentator (bottom) test sets, reported with 95\% confidence intervals. This table presents the performance of FedMust using only 60 labeled cases per client under two scenarios: without using local unlabeled data (FedMust) and with local unlabeled data incorporated during student training (FedMust+UL).}\label{table_use_ulbl_60}
    \resizebox{\textwidth}{!}{%
    \begin{tabular}{l l *{3}{c}}
        \toprule
        & & \multicolumn{3}{c}{\textbf{Ensemble Dice Score}} \\
        \cmidrule(lr){3-5}
        \textbf{Testset} & \textbf{Task} &
        \textbf{Local} & \textbf{FedMust} & \textbf{FedMust+UL} \\
        \midrule
        \multirow{4}{*}{AMOS22}      & Liver      & \textbf{0.863}~(0.848,0.876) & 0.846~(0.828,0.864) & 0.842~(0.825,0.859) \\ 
                                   &  Kidney     & 0.537~(0.492,0.581) & 0.642~(0.601,0.681) & \textbf{0.677}~(0.639,0.713) \\ 
                                   & Pancreas   & 0.282~(0.257,0.307) & \textbf{0.324}~(0.295,0.353) & 0.312~(0.282,0.342) \\
                                   &  Spleen     & 0.630~(0.588,0.670) & \textbf{0.668}~(0.628,0.707) & 0.652~(0.611,0.691) \\ 
        \midrule
        \multirow{4}{*}{TotalSeg}      & Liver      & 0.869~(0.856,0.882) & \textbf{0.873}~(0.860,0.886) & 0.859~(0.842,0.874) \\ 
                                     &  Kidney     & 0.570~(0.540,0.600) & 0.573~(0.544,0.603) & \textbf{0.574}~(0.545,0.603) \\ 
                                     & Pancreas   & 0.330~(0.310,0.349) & 0.408~(0.387,0.429) & \textbf{0.412}~(0.390,0.434) \\
                                     &  Spleen     & 0.573~(0.545,0.601) & \textbf{0.679}~(0.652,0.705) &  0.631~(0.603,0.659) \\ 
        \bottomrule
    \end{tabular}}
\end{table*}

In Section~\ref{sec_num_lbl}, only the available annotated data at each client were used to generate features for all tasks. Subsequently, only this portion of the data was employed during student training. However, in real-world settings, institutions possess a limited amount of annotated data alongside a substantially larger amount of unlabeled data that often remains unused. To examine the potential of leveraging such data, we design this experiment such that each client has partial labels for only 60 cases from its local dataset, while the remaining cases are treated as unlabeled. Consequently, during teacher training, only 60 cases are used for tasks with annotations at each client; however, during student training, features are generated and utilized for the entire available dataset, including both labeled and unlabeled cases. For example, for client C1, as shown in Table~\ref{annotation_table}, 60 cases are used to train the teacher model, after which the teacher models from all clients are used to generate features for all 131 cases across all tasks.

The Dice performance of the models trained in this manner is then compared with that of the local models and the corresponding scenario in Section~\ref{sec_num_lbl}. The ensembled performance of the 5-fold cross-validated models and the corresponding confidence intervals, evaluated on the AMOS and TotalSegmentator test sets, are summarized in Table~\ref{table_use_ulbl_60}. The results show that incorporating additional unlabeled data from the clients leads to inconsistent performance improvements compared with FedMust. We hypothesize that this inconsistent behavior stems from unreliable teachers that fail to provide meaningful features for the unlabeled data during the early stages of training. This behavior could potentially be mitigated by using pretrained models or extending the initial teacher FL round to start the process with more reliable teachers. The same inconsistent behavior is also observed in the experiment with 20 labeled cases per client, as opposed to the 60 cases considered here, with the corresponding results provided in the Supplementary Material.

\subsection{Federated Participation Without Annotations}\label{unlabled_cli_sec}
The traditional FL approach, as well as its recent semi-supervised variants, requires clients to have labeled data for at least one task to participate in the federation. FedMust, however, relaxes this requirement by allowing clients without labeled data to participate in FL and potentially contribute to improved federated performance by utilizing their unlabeled data during student training. To investigate this, we consider a scenario with five clients, four of which are identical to those in the previous setups and use all their available labeled data, while the fifth client uses cases from the TotalSegmentator test set as unlabeled. Consequently, the fifth client participates in the federation exclusively during student training, where the teacher models are used to generate features from its local data, which are subsequently utilized for student training. 

\begin{table*}[t]
    \centering
    \caption{Ensembled Dice performance of models trained using five-fold cross-validation and evaluated on the independent AMOS test set, reported with 95\% confidence intervals. FedMust-4C denotes the setting with four clients, each having annotations for a single task. FedMust-5C uses the same four clients with an additional fifth client that has no labeled data and contributes only unlabeled data from the TotalSegmentator dataset.}\label{table_5th_client}
    \resizebox{\textwidth}{!}{%
    \begin{tabular}{l l *{3}{c}}
        \toprule
        & & \multicolumn{3}{c}{\textbf{Ensemble Dice Score}} \\
        \cmidrule(lr){3-5}
        \textbf{Testset} & \textbf{Task} &
        \textbf{Local} & \textbf{FedMust-4C} & \textbf{FedMust-5C} \\
        \midrule
        \multirow{4}{*}{AMOS22}      & Liver      & 0.865~(0.849,0.879) & \textbf{0.878}~(0.864,0.892) & 0.871~(0.856,0.886) \\ 
                                   &  Kidney     & 0.784~(0.754,0.812) & \textbf{0.835}~(0.813,0.855) & 0.830~(0.808,0.851) \\ 
                                   & Pancreas   & 0.445~(0.412,0.477) & \textbf{0.519}~(0.486,0.551) & 0.484~(0.450,0.518) \\
                                   &  Spleen     & 0.630 (0.588,0.670) & \textbf{0.681}~(0.641,0.719) & 0.660~(0.619,0.700) \\ 
        \bottomrule
    \end{tabular}}
\end{table*}

\begin{figure}[t]
    \centering
    \includegraphics[width=\textwidth]{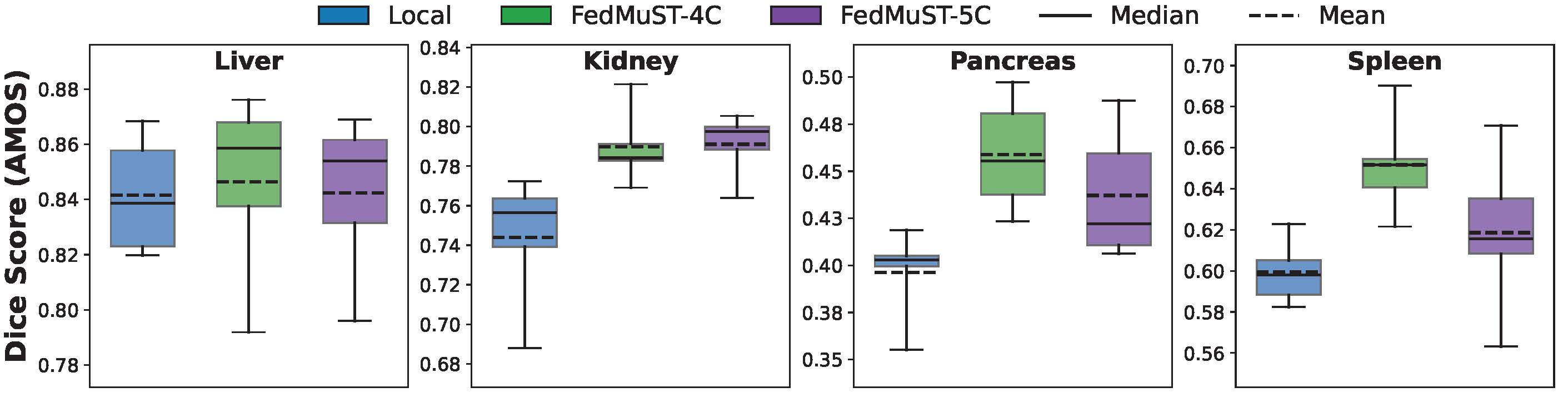}\vspace{-0.5mm}
    \caption{Model performance across the five cross-validation folds, evaluated on the AMOS test set. The results show the performance of individual folds rather than the ensembled performance. FedMust-4C denotes the setting with four clients, each having annotations for a single task. FedMust-5C uses the same four clients with an additional fifth client that has no labeled data and contributes only unlabeled data from the TotalSegmentator dataset.}\label{5th_client_fig}
\end{figure}

The trained model is then evaluated by comparing its Dice performance with that of the local model and the corresponding FedMust model using four clients. This comparison isolates the effect of incorporating an unlabeled client into the federation. The ensembled performance of the 5-fold cross-validated models on the AMOS test set, along with the corresponding confidence intervals, is summarized in Table~\ref{table_5th_client}. The models are evaluated only on the AMOS test set, since the data from the fifth client were obtained from the TotalSegmentator cohort, and comparisons with the four-client scenario would not provide a fair comparison due to data leakage. The results indicate that adding a fifth client to the federation does not improve the performance of the existing FL clients, suggesting that augmenting FL training with additional unlabeled data is not necessarily beneficial. This is also shown in the box plots in Fig.~\ref{5th_client_fig}, which illustrate model performance across the five cross-validation folds on the AMOS test set. These results further confirm that the inclusion of a fifth client with a large amount of unlabeled data in the FL process does not yield consistent performance improvements compared with the four-client FL scenario. Such inconsistent behavior may be attributed to the imbalance between the amount of unlabeled data available at the fifth client and the data sizes of the other clients.

\section{Discussion}\label{sec_discu}
The impact of unlabeled data, whether originating from other clients or from the client itself, on the performance of FedMust requires further clarification. Since the data across clients are only partially labeled, in this scenario, data with annotations for a specific task are considered unlabeled for other clients. For example, when generating features using the Liver teacher, data from C2, C3, and C4 are considered unlabeled, as C1 is the only client with labeled data for the liver task. As shown in Section~\ref{sec_num_lbl}, incorporating such unlabeled data from other clients during student training consistently improves segmentation performance for each task. However, Section~\ref{local_unlabled_sec} shows that, when the contribution of unlabeled data from other clients is fixed, additionally incorporating unlabeled data from the clients themselves results in inconsistent performance improvements. A possible explanation for these marginal performance improvements is that leveraging additional local unlabeled data may be beneficial only when the teacher model used to generate the features is capable of providing meaningful feature representations in the first place. To address this limitation, future experiments could initialize the FL process using pretrained models or extend the initial federated round with more local iterations to obtain better-performing teacher models. In addition, the use of foundation models or more sophisticated teacher models that can maintain high performance under limited labeled data requires further investigation in future work.

Unlike the methods proposed in the literature, FedMust allows clients with no labeled data to participate in the federation, as investigated in Section~\ref{unlabled_cli_sec}. The results show that including an unlabeled client with substantially more data than the labeled data available at the other clients is not, overall, beneficial compared with the four-client setting. However, introducing such an unbalanced amount of unlabeled data into the federation, even under the extreme scenario in which clients have no overlapping sets of labeled tasks, results in a stable trained model with improved performance compared with models trained in isolation. These findings motivate further investigation under more realistic scenarios involving a larger number of clients, overlapping sets of labeled tasks, and a more balanced data distribution. An additional consideration is that FedMust introduces  computational overhead due to the training of the student encoder compared with traditional FL approaches. However, given recent advances in computational hardware and the increasing availability of GPUs, this additional computational requirement is not expected to be a major concern.

\section{Conclusion}\label{sec_conclu}

This study proposed a semi-supervised federated learning (FL) framework for multi-organ segmentation that, unlike conventional FL algorithms, handles inconsistent label availability across clients. The proposed approach adopted a multi-task student–teacher framework to leverage both labeled and unlabeled data to enhance segmentation performance. The results demonstrated that the proposed framework outperformed the baseline models by leveraging multi-task learning and unlabeled data from other clients. However, incorporating additional unlabeled data from the clients themselves was not consistently beneficial. While the proposed method relaxed the requirement for clients to possess labeled data to participate in FL, the results did not favor the inclusion of an additional client with a large amount of unlabeled data in the federation.

\section*{Declarations}
\textbf{Funding.} Funding: Project FLIP.AI: Federated Learning to Improve Prostate cancer imaging with Artificial Intelligence, funding from Norwegian Cancer Society (project code:~215951), Norwegian University of Science and Technology, Department of Circulation and Medical Imaging.

\noindent\textbf{Conflict of/Competing interests.} The authors declare no conflicts of interest. 

\noindent \textbf{Data availability.} All participant information and data used in this study have been previously published, publicly available, and are appropriately cited in the manuscript (data from the Medical Segmentation Decathlon challenge~\cite{antonelli2022medical}, the KiTS19 challenge~\cite{heller2019kits19}, AMOS~\cite{ji2022amos}, and TotalSegmentator~\cite{wasserthal2023totalsegmentator}). 

\noindent \textbf{Code availability.} All code and links to the data are shared via GitHub and are publicly available at https://github.com/AshknMrd/FedMust. 

\bibliography{Ref_arxiv}

\end{document}